\documentclass{article}

\usepackage[preprint]{neurips_2026}

\usepackage{microtype}      
\usepackage[utf8]{inputenc} 
\usepackage[T1]{fontenc}    
\usepackage{hyperref}       
\usepackage{url}            
\usepackage{booktabs}       
\usepackage{amsfonts}       
\usepackage{nicefrac}       
\usepackage{xcolor}         
\usepackage{algorithm}
\usepackage{amsmath}
\usepackage{makecell}
\usepackage{multirow}
\usepackage{algpseudocode}
\usepackage{amssymb}
\usepackage{ragged2e}
\usepackage{graphicx}
\usepackage{svg}
\usepackage{float}
\usepackage{algorithmicx}
\usepackage{tikz}
\usepackage{caption}
\usepackage{amsthm}
\theoremstyle{definition}
\newtheorem{definition}{Definition}[section]

\title{VDGR-RAG: Vectors, Directories, Graphs, and Reflection Are All You Need for Unified Reasoning over Hierarchical Enterprise Knowledge}

\author{%
  Wenqi Chen, Haofei Yang, Rui Yang, Fangming Li \\
  ICT AI Competence Center, Huawei Technologies, Shanghai, China \\
  \texttt{\{chenwenqi10, yangrui235, yanghaofei2, lifangming1\}@huawei.com} \\
  Corresponding authors: Rui Yang, Fangming Li
}

\begin{document}

\maketitle

\begin{abstract}
Retrieval-Augmented Generation (RAG) is essential for enterprise knowledge question answering (QA), particularly in domains with complex product documentation like telecommunications. However, existing RAG approaches largely overlook the holistic integration of diverse retrieval strengths, leading to inaccurate domain routing, poor utilization of hierarchical document structures, and consequently limited reasoning capabilities over enterprise knowledge. To address these limitations, we present VDGR-RAG, which integrates vector retrieval, directory-driven reasoning, graph traversal, and iterative reflection in a unified framework for accurate enterprise knowledge QA. Specifically, VDGR-RAG is an agentic GraphRAG system that first constructs a Hierarchical Heterogeneous Knowledge Graph ($\text{H}^2$KG) from document chunks to preserve both hierarchical directory structures and semantic relationships, and then employs a set of atomic tools for knowledge retrieval that can be freely composed to navigate the $\text{H}^2$KG: (1) a directory-enhanced routing tool that uses table-of-contents (TOC) structures to route user queries to appropriate domain-specific $\text{H}^2$KGs; (2) a multi-route retrieval tool that combines vector search, TOC-based agentic search, and graph search for comprehensive knowledge retrieval; (3) a directory backtracking tool that corrects knowledge localization biases; and (4) a dynamic reflection tool that iteratively plans the next retrieval phase. We conduct extensive experiments on our enterprise product documents across four wireless domains (e.g., energy saving and fault management). Experimental results demonstrate that our method significantly outperforms a variety of RAG baselines in terms of both knowledge retrieval recall and QA accuracy.
\end{abstract}

\section{Introduction}

Large Language Models (LLMs), such as GPT-5.4~\citep{2026From} and Qwen-3~\citep{2025Qwen3}, have revolutionized enterprise knowledge management and question answering (QA) systems~\citep{Yue2025A,2023Complex}. Driven by these advancements, organizations increasingly rely on LLMs to automate knowledge-intensive tasks, such as technical support~\citep{2026Patent}, product documentation retrieval~\citep{2026Context}, and operational guidance~\citep{2025Bridging}. However, LLMs face critical limitations: knowledge cutoff constraints and hallucination risks often lead to outdated or factually incorrect responses~\citep{2022SurveyHallucination}. To address these issues, Retrieval-Augmented Generation (RAG) has emerged as the dominant paradigm~\citep{Mei2025A, fan2024survey, 2024Graph, li2025survey}, retrieving relevant knowledge from external corpora to augment the model context and improve QA accuracy.

In practical enterprise applications, especially within complex domains like telecommunications, applying RAG to technical documentation presents several critical challenges. First, these documents exhibit pervasive lexical and conceptual overlap across standard specifications. For instance, 4G LTE and 5G NR share highly similar technical descriptions for analogous parameters and procedures, making it difficult for retrieval models to distinguish relevant passages from superficially similar yet contextually distinct ones. Second, telecommunications documents possess an inherent multi-level hierarchical structure with extensive cross-section and cross-document references. Conventional RAG paradigms, which treat documents as flat text chunks or build knowledge graphs for the open domain, cannot exploit this structural richness to simultaneously resolve domain-specific terminological ambiguity and preserve multi-level structural topology~\citep{2025BookRAG}. These limitations severely hinder accurate reasoning performance over complex enterprise knowledge bases.

Existing RAG approaches for enterprise document QA can be broadly divided into three categories. The first category, \textit{vector-based RAG}, divides documents into text chunks and constructs vector indices for similarity search~\citep{2020RetrievalAugmented, 2023RetrievalAugmented}, offering computational efficiency and ease of deployment, making it well-suited for enterprise applications. The second category, \textit{graph-based RAG}, constructs structured relationships between text units to model explicit document topology. Representative paradigms like GraphRAG~\citep{2024From} and RAPTOR~\citep{2024RAPTORRecursive} build entity-level knowledge graphs or recursive document trees, while recent advances such as HippoRAG~\citep{Guti2024HippoRAG} and NodeRAG~\citep{2025NodeRAG} leverage heterogeneous graph structures to enable cross-chunk relational retrieval. 
The third category, \textit{agentic RAG}, or reasoning-based RAG, employs LLM-driven autonomous search and page-level index navigation (e.g., PageIndex~\citep{zhang2025pageindex}) to dynamically reason and reflect over document directories and iteratively retrieve relevant content. Importantly, graph-based RAG and agentic RAG are not mutually exclusive; rather, they upgrade traditional vector-based RAG along two orthogonal dimensions—the index level and the retrieval level, respectively (Section 2). Despite these advancements, existing methods suffer from two major limitations:
\begin{itemize}
    \item \textbf{Failure to synergize heterogeneous retrieval paradigms.} Current RAG systems struggle to fully capitalize on the complementary strengths of the above three paradigms. Vector-based RAG offers low computational latency, but treating documents as flat chunks severs global semantic coherence and loses structural context~\citep{liu2024lost}. Furthermore, the discrepancy between semantic similarity and true relevance often leads to high false-positive rates~\citep{zhang2025pageindex}. Agentic RAG achieves higher precision through dynamic navigation, yet incurs excessive latency over large-scale enterprise corpora. Meanwhile, many graph-based RAG approaches primarily target open-domain text. Relying on generic NLP toolkits (e.g., spaCy in LinearRAG~\citep{2025LinearRAG}) fails to extract specialized terminology and non-trivial relations in technical domains like telecommunications. Crucially, none of the existing frameworks effectively integrate the complementary strengths of these distinct retrieval paradigms over structured enterprise knowledge.
    \item \textbf{Absence of routing across domain knowledge bases.} Enterprise documentation spans multiple specialized knowledge bases with overlapping terminology (e.g., distinct databases of energy saving and fault management for wireless devices). Without an effective domain-level routing mechanism, current paradigms resort to exhaustive searches across all knowledge bases, which introduces severe context noise and significantly increases query latency.
\end{itemize}

To address these limitations, we argue that vector representation, directory structures, graph topology, and dynamic reflection must be deeply synergized to effectively reason over complex hierarchical enterprise knowledge bases. To this end, we propose VDGR-RAG, an agentic GraphRAG framework that integrates vector retrieval, directory-driven reasoning, graph traversal, and iterative reflection into a unified system for accurate enterprise knowledge QA. Specifically, VDGR-RAG first constructs a Hierarchical Heterogeneous Knowledge Graph ($\text{H}^2$KG) from the document corpus to preserve both hierarchical directory structures and semantic entity relationships, and then deploys a suite of composable, atomic tools for knowledge search. Rather than enforcing a rigid, linear search pipeline, our approach empowers the LLM agent to dynamically coordinate these tools based on intermediate retrieval feedback to navigate the $\text{H}^2$KG. The toolset of VDGR-RAG is specified as follows:
\begin{itemize}
    \item \textbf{Directory-Enhanced Routing Tool:} To tackle the severe context noise caused by exhaustive cross-domain searches, this tool leverages domain knowledge and multi-level TOC structures as contextual grounding. By projecting user queries onto candidate domain TOC trees, it generates multi-step routing decisions with confidence scores, precisely steering queries to target domain knowledge bases while pruning irrelevant ones.
    
    \item \textbf{Multi-Route Retrieval Tool:} To integrate the complementary strengths of diverse retrieval modes, this tool orchestrates three heterogeneous pathways: (i) \textit{naive RAG} that combines dense vector search with BM25 keyword matching for rapid hybrid retrieval, (ii) \textit{TOC-based agentic search} that leverages vector-based coarse filtering over section titles followed by LLM-driven fine selection for structure-aware navigation, and (iii) \textit{graph search} that extracts key query concepts as seeds to execute Personalized PageRank (PPR) over the knowledge graph, enabling single-step multi-hop reasoning and chunk-entity subgraph discovery. This multi-route retrieval guarantees high knowledge recall without losing structural context.
    
    \item \textbf{Directory Backtracking Tool:} When the initial multi-route retrieval yields insufficient context or suffers from localization bias, this tool enables the LLM agent to explore neighboring regions of the document hierarchy starting from the currently retrieved position. By analyzing parent section summaries and structural information, it dynamically traverses upward and laterally searches across the TOC tree to retrieve missing prerequisite context and ensure complete knowledge coverage.
    
    \item \textbf{Dynamic Reflection Tool:} Serving as the brain of the agentic workflow, this tool continuously evaluates the sufficiency and relevance of retrieved contexts. Upon detecting ambiguity, information gaps, or execution failures, it autonomously plans the next search query until a complete, high-confidence answer can be generated.
\end{itemize}

The remainder of this paper is organized as follows. Section 2 reviews related work on existing RAG techniques. Section 3 introduces preliminary knowledge. Section 4 presents the methodology of our proposed Hierarchical Heterogeneous Knowledge Graph ($\text{H}^2$KG) construction. Section 5 details the retrieval methods of VDGR-RAG. Section 6 presents the experimental evaluations and performance analysis. Finally, Section 7 concludes the paper.

\section{Related Work}
In this section, we review the evolution of Retrieval-Augmented Generation (RAG) paradigms across three key stages: standard RAG systems, index-level enhancements through graph-based RAG, and retrieval-level upgrades via Agentic RAG.

\subsection{Standard RAG Systems}
Retrieval-Augmented Generation (RAG) was introduced by Lewis et al.~\citep{2020RetrievalAugmented} to reduce LLM hallucinations by integrating external knowledge. Standard RAG pipelines typically follow a three-stage workflow: chunking, embedding-based retrieval, and LLM generation. Early sparse and dense retrieval baselines relied on BM25~\citep{2007Theprobabilistic} and Dense Passage Retrieval (DPR)~\citep{2020DensePassage}. Modern systems primarily deploy vector search via pre-trained encoders, such as BGE-Embedding~\citep{2023bge} and Qwen-Embedding~\citep{2025Qwen3Embedding}, using approximate nearest neighbor algorithms like HNSW~\citep{Yury2018Efficient} for semantic matching. However, these baseline methods treat documents as flat text collections. By relying solely on point-to-point term or vector similarity, they ignore the logical hierarchy of technical documents and often miss critical contextual information.

\subsection{Graph-based RAG: Index-Level Enhancements}
To overcome the limits of flat retrieval, recent works focus on index-level upgrades by introducing knowledge graphs. GraphRAG~\citep{2024From} constructs entity graphs and uses community detection to provide global context for queries. RAPTOR~\citep{2024RAPTORRecursive} builds recursive trees through iterative clustering and summarization to support multi-level semantic understanding. Many other efforts, such as HippoRAG~\citep{Guti2024HippoRAG}, NodeRAG~\citep{2025NodeRAG}, SubgraphRAG~\citep{li2025simple}, and E$^2$GraphRAG~\citep{2025E2GraphRAG}, leverage graph topologies to model heterogeneous document relationships. By building internal connections within texts, these methods shift retrieval from simple point-to-point matching to structured context understanding. However, most of these graph-based RAG approaches focus heavily on graph entity links while largely ignoring inherent document hierarchies, such as multi-level tables-of-contents (TOC). Some works have attempted to leverage these hierarchies by introducing layout-aware parsers~\citep{2020LayoutLM, 2021LayoutParser} or constructing document tree index structures~\citep{2025BookRAG, niuetal2025tree, zhang2025pageindex}. However, these approaches struggle to leverage the distinct advantages of TOC structures and graph entity topologies. Although recent efforts like BookRAG~\citep{2025BookRAG} attempt to combine both, its TOC-based reasoning is strictly limited to mining additional entities to enrich the entity subgraph where PPR operates, rather than guiding direct knowledge search. Crucially, existing approaches lack a domain routing mechanism that exploits TOC information across vast, multi-document repositories. Without explicit domain routing and structural integration, these systems resort to exhaustive global searches across all technical manuals, introducing severe context noise and significantly increasing query latency.

\subsection{Agentic RAG: Retrieval-Level Upgrades}
Beyond index structures, Agentic RAG shifts attention to upgrading the retrieval workflow itself by using LLMs as autonomous agents~\citep{2022ReAct}. Self-RAG~\citep{asai2023selfraglearningretrievegenerate} enables models to self-critique and refine retrieved content. Other agent-based frameworks~\citep{trivedi2023, zhang2025pageindex, 2026ARAG, 2025BookRAG, 2024Adaptive, singh2026agenticretrievalaugmentedgenerationsurvey} deploy LLMs to plan search trajectories, decompose complex queries, and select suitable retrieval tools. By introducing LLM-driven decision-making and dynamic reflection, Agentic RAG transforms static lookup into interactive, multi-step deep search. Despite these retrieval-level advances, current agentic methods lack a unified framework that seamlessly combines hybrid vector search, TOC-based tree navigation, graph traversal, and dynamic reflection for complex reasoning on enterprise knowledge.

\section{Preliminaries}

In this section, we formally define the enterprise knowledge QA task, introduce the general RAG paradigm, and describe the naive RAG baseline that serves as the foundation of many QA systems.

\subsection{Enterprise Knowledge QA Task}

Given a collection of technical documents $\mathcal{D} = \{d_1, d_2, \ldots, d_n\}$ and a user query $q$, the goal of enterprise knowledge QA is to retrieve relevant evidence $\mathcal{E}$ from $\mathcal{D}$ and generate an accurate answer $a$ that faithfully reflects the retrieved evidence. Formally:
\begin{equation}
	a = \text{Generate}(q, \mathcal{E}), \quad \mathcal{E} \subseteq \text{Index}(\mathcal{D})
\end{equation}
where $\text{Index}(\mathcal{D})$ denotes the indexed representation of the document corpus, and $\text{Generate}(q, \mathcal{E})$ represents the LLM-based answer synthesis process. The key challenge lies in ensuring that $\mathcal{E}$ comprehensively covers all necessary evidence for $q$, especially when relevant knowledge is dispersed across multiple documents, sections, and hierarchical levels within complex enterprise documentation.

\subsection{RAG Paradigm: Offline Indexing and Online Retrieval}

The RAG paradigm consists of two sequential phases: offline indexing and online retrieval-generation.

\textbf{Offline Indexing.} The document corpus $\mathcal{D}$ is preprocessed into retrievable units through a two-stage pipeline. In the first stage, documents are parsed and partitioned into a chunk set $\mathcal{C}=\{c_1, c_2, \ldots, c_m\}$, where each chunk $c = \{\texttt{Content}, \texttt{Source}, \texttt{Title}\}$ consists of three fields: \texttt{Content} (the raw text of the chunk), \texttt{Source} (the file path locating the chunk within the document hierarchy), and \texttt{Title} (the section heading under which the chunk appears). These chunks are indexed for both sparse and dense retrieval, forming the base knowledge pack. In the second stage, based on these chunks, a Hierarchical Heterogeneous Knowledge Graph ($\text{H}^2$KG) $\mathcal{G}$ is constructed to capture hierarchical directory structures and semantic relationships among chunks, entities, and document sections. Together, the chunk-level index and the $\text{H}^2$KG constitute the complete offline knowledge base $\mathcal{K}$.

\textbf{Online Retrieval and Generation.} Given a user query $q$, the retrieval module searches the indexed knowledge base $\mathcal{K}$ to collect relevant chunks, which are then fed into the LLM for answer generation. The quality of the final answer $a$ depends critically on the comprehensiveness and relevance of the retrieved evidence $\mathcal{E}$.

\subsection{Naive RAG: Hybrid Retrieval and Re-ranking}

As the most basic instantiation of the RAG paradigm, naive RAG primarily relies on dense vector retrieval for semantic matching, augmented with sparse lexical search (BM25) and neural re-ranking.

\textbf{Sparse and Dense Candidate Retrieval.} Sparse retrieval (BM25 via Elasticsearch) computes lexical matching scores over the textual content of each chunk and returns the top candidates by lexical relevance. Simultaneously, dense vector retrieval calculates the cosine similarity between the query and chunk embeddings in the semantic space and returns the top candidates by semantic relevance. The results from both pathways are then merged via score fusion (e.g., Reciprocal Rank Fusion~\citep{cormack2009reciprocal}) to form a unified candidate set.

\textbf{Score Fusion and Re-ranking.} A cross-encoder re-ranker computes final relevance scores over the merged candidate set to select the top-$K$ chunks:
\begin{align}
	\mathcal{H} &= \text{HybridSearch}(q, \mathcal{C}) \nonumber \\
	\mathcal{R} &= \text{ReRank}(\mathcal{H}) \nonumber \\
	\mathcal{E}_{\text{Vanilla}} &= \text{TopK}(\mathcal{R}, K)
\end{align}
where $\mathcal{H}$ denotes the merged candidate set from score fusion of sparse and dense retrieval over query $q$ and the raw chunk set $\mathcal{C}$, $\mathcal{R}$ denotes the re-ranked set after cross-encoder re-scoring, and $\mathcal{E}_{\text{Vanilla}}$ denotes the final top-$K$ chunks selected for answer generation.
While computationally efficient and easy to deploy, this naive approach treats documents as flat text collections without leveraging their hierarchical structures or semantic relationships, which limits its effectiveness on complex enterprise knowledge bases.

\section{\texorpdfstring{Hierarchical Heterogeneous Knowledge Graph ($\text{H}^2$KG) Construction}{Hierarchical Heterogeneous Knowledge Graph (H2KG) Construction}}
\begin{figure*}[!t]
	\centering
	\includegraphics[width=\textwidth]{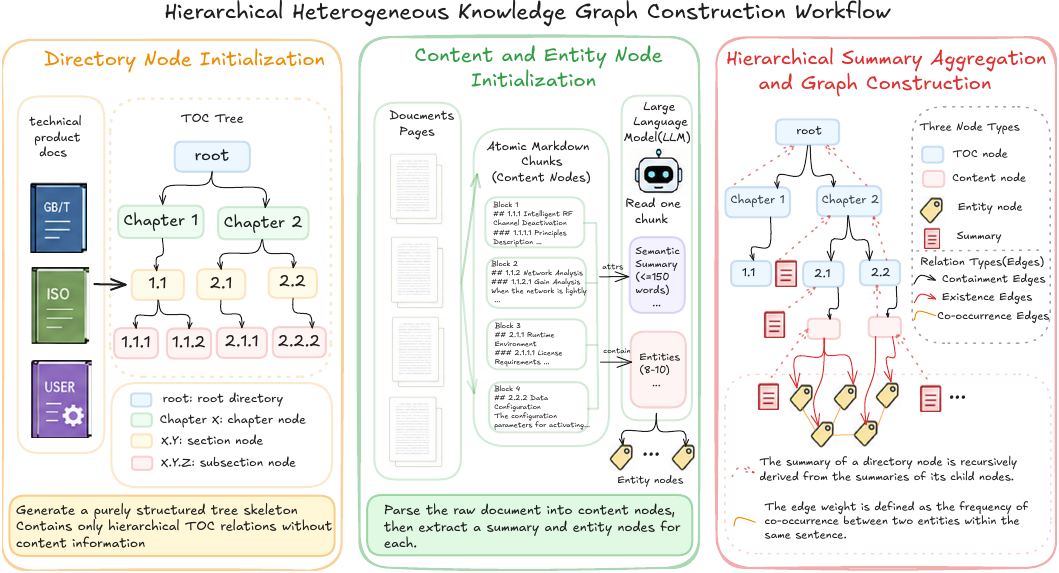} 
	\caption{Overview of the $\text{H}^2$KG construction workflow. (Left) Directory nodes are extracted from document TOC hierarchies. (Middle) Content nodes correspond to chunks, with entity nodes and chunk summaries extracted by an LLM from each chunk. (Right) The final $\text{H}^2$KG integrates directory nodes, content nodes, and entity nodes with three types of edges (containment, existence, and co-occurrence), and generates directory summaries in a bottom-up manner for each directory node.}
	\label{fig:graph_cropped}
\end{figure*}

This section details the construction of the Hierarchical Heterogeneous Knowledge Graph ($\text{H}^2$KG). As introduced in Section 3, the offline indexing pipeline first parses raw documents into text chunks, and then constructs the $\text{H}^2$KG over these chunks to preserve both hierarchical directory structures and entity relationships. The overall construction workflow is illustrated in Figure~\ref{fig:graph_cropped}. In this section, we first establish the formal topology and definition of the $\text{H}^2$KG. Next, we describe node initialization from the chunk set. We then present edge construction and LLM-driven semantic augmentation. Finally, we detail the global conflation procedure.

\subsection{\texorpdfstring{Overview of the $\text{H}^2$KG Structure}{Overview of the H2KG Structure}}
Complex technical documentation requires a hybrid knowledge representation that bridges top-down document hierarchies with bottom-up semantic topologies. To fulfill this need, we design the $\text{H}^2$KG as a unified structural-semantic graph.

\begin{definition}[Hierarchical Heterogeneous Knowledge Graph]
The $\text{H}^2$KG is defined as a heterogeneous graph $\mathcal{G}=(\mathcal{V},\mathcal{E},\tau,\phi)$, where:
\begin{itemize}
	\item $\mathcal{V}=\mathcal{V}_\mathcal{D}\cup\mathcal{V}_\mathcal{C}\cup\mathcal{V}_\mathcal{K}$ represents the node set partitioned into three categories:
	\begin{itemize}
		\item $\mathcal{V}_\mathcal{D}$ (\textbf{Directory Nodes}): Nodes parsed from the native table-of-contents (TOC) hierarchy of the original documents.
	    \item $\mathcal{V}_\mathcal{C}$ (\textbf{Content Nodes}): Atomic semantic units of text retrieval, each corresponding to a chunk from the parsed document.
	    \item $\mathcal{V}_\mathcal{K}$ (\textbf{Entity Nodes}): Fine-grained technical entities extracted from chunks, such as domain concepts and specialized terminology.
	\end{itemize}
	\item $\mathcal{E}=\mathcal{E}_\supset\cup\mathcal{E}_\exists\cup\mathcal{E}_{\mathrm{cooccur}}$ represents the edge set partitioned into three relationship types:
	\begin{itemize}
		\item $\mathcal{E}_\supset\subseteq\mathcal{V}_\mathcal{D}\times(\mathcal{V}_\mathcal{D}\cup\mathcal{V}_\mathcal{C})$ (\textbf{Containment Edges}): Directed edges from parent nodes to their child nodes, capturing the document hierarchy.
	    \item $\mathcal{E}_\exists\subseteq\mathcal{V}_\mathcal{C}\times\mathcal{V}_\mathcal{K}$ (\textbf{Existence Edges}): Directed edges from content to entity, indicating that an entity explicitly occurs in the content block.
		\item $\mathcal{E}_{\mathrm{cooccur}}\subseteq\mathcal{V}_\mathcal{K}\times\mathcal{V}_\mathcal{K}$ (\textbf{Co-occurrence Edges}): Semantic links between entities (i.e., two entities co-occur in the same sentence, inspired by~\citep{2025E2GraphRAG}).
	\end{itemize}
	\item $\tau:\mathcal{V}\rightarrow\{D,C,K\}$ is the node type-mapping function that assigns each node to its functional category: directory ($D$), content ($C$), or entity ($K$).
	\item $\phi:\mathcal{E}\rightarrow\{\supset,\exists,\mathrm{cooccur}\}$ is the edge type-mapping function that assigns each edge to its relationship type: containment ($\supset$), existence ($\exists$), or co-occurrence ($\mathrm{cooccur}$).
\end{itemize}
\end{definition}

\subsection{Node Initialization from Chunk Set}

Given the chunk set $\mathcal{C}$ produced by the offline indexing pipeline (Section 3), we initialize three types of nodes for the $\text{H}^2$KG: directory nodes, content nodes, and entity nodes.

\noindent\textbf{Directory Nodes.} As shown on the left side of Figure~\ref{fig:graph_cropped}, directory nodes are extracted from the document's table-of-contents (TOC) hierarchy, including the document root node, chapter nodes, section nodes, and all subsection nodes. Each directory node $v_d\in\mathcal{V}_\mathcal{D}$ is initialized with the following attributes: $\texttt{Path}_{\texttt{full}}$, $\texttt{Path}_{\texttt{parent}}$, and $\texttt{Ver}_{\texttt{prod}}$, where $\texttt{Path}_{\texttt{full}}$ denotes the fully concatenated path string from the document root to the current section, $\texttt{Path}_{\texttt{parent}}$ denotes the path of the immediate parent section (which is used for building containment edges and enabling directory backtracking), and $\texttt{Ver}_{\texttt{prod}}$ is a root-level attribute that tracks product version constraints, allowing the retrieval phase to filter out chunks from irrelevant product versions.

\noindent\textbf{Content Nodes.} Each chunk $c \in \mathcal{C}$ maps directly to a content node $v_c\in\mathcal{V}_\mathcal{C}$, as shown in the middle panel of Figure~\ref{fig:graph_cropped}. To ensure unique tracking, the primary identifier of $v_c$ is derived from its raw text content combined with its canonical file path. At initialization, content nodes carry the following attributes: the raw text content (\texttt{Content}) and structural metadata (\texttt{Source} and \texttt{Title}) inherited from the chunk set.

\noindent\textbf{Entity Nodes.} As shown in the middle panel of Figure~\ref{fig:graph_cropped}, entity nodes $v_k\in\mathcal{V}_\mathcal{K}$ are fine-grained technical entities extracted from content chunks by an LLM, including technical terms, domain concepts, device names, and other specialized terminology. Each entity node is assigned two attributes: $\texttt{Name}$, which is the output of the LLM extraction process, and an $\texttt{Importance Score}$ $\lambda\in[1,10]$ simultaneously output by the LLM during extraction, reflecting the entity's importance to the corresponding content node. To ensure high-quality entity extraction from telecommunications documents, the LLM extraction enforces three criteria: (i) \textit{Relevance}: prioritize operational concepts and domain-specific parameters over generic terms; (ii) \textit{Distinctiveness}: exclude common filler words that lack discriminative power; and (iii) \textit{Exclusion}: omit volatile numerical literals and low-level internal identifiers that vary across product versions.

\subsection{Edge Construction}

After node initialization, we construct three types of edges to establish structure-guided retrieval paths, as shown on the right side of Figure~\ref{fig:graph_cropped}.

\noindent\textbf{Containment Edges ($\mathcal{E}_\supset$).} Directed containment edges capture the document hierarchy from parent nodes to their children:
\begin{itemize}
	\item \textbf{Directory-to-Directory:} An edge $e\in\mathcal{E}_\supset$ connects $v_{d_i}$ to $v_{d_j}$ if $v_{d_i}$ is the direct parent of $v_{d_j}$ in the TOC hierarchy, which can be efficiently determined by checking whether $\mathrm{Path}_{\mathrm{full}}(v_{d_i})$ equals $\mathrm{Path}_{\mathrm{parent}}(v_{d_j})$, the full path of $v_{d_j}$'s parent directory.
	\item \textbf{Directory-to-Content:} An edge connects a leaf directory node to each content node $v_c$ that belongs to it.
\end{itemize}

\noindent\textbf{Existence Edges ($\mathcal{E}_\exists$).} Directed existence edges connect content nodes to entity nodes, indicating that the content block contains knowledge about the relevant entities. When the LLM extracts entities from a content node $v_c$ (Section 4.2), a directed edge $e=(v_c,v_k)\in\mathcal{E}_\exists$ is instantiated. This mapping allows graph traversal from technical entities to their underlying source text.

\noindent\textbf{Co-occurrence Edges ($\mathcal{E}_{\mathrm{cooccur}}$).} Inspired by E$^2$GraphRAG~\citep{2025E2GraphRAG} and LinearRAG~\citep{2025LinearRAG}, we deliberately avoid using LLMs to extract explicit relations between entities, as relation triple extraction is computationally expensive and introduces excessive noise due to the inaccuracies and uncertainties of LLM-generated relations. Instead, we adopt a frequency-based co-occurrence approach to establish semantic associations between entities. Specifically, the textual contents are split into discrete sentences, and an undirected edge is created between entities $v_{k_i}\in\mathcal{V}_\mathcal{K}$ and $v_{k_j}\in\mathcal{V}_\mathcal{K}$ if they co-occur within the same sentence. The edge weight $w(v_{k_i}, v_{k_j})$ is assigned as the raw co-occurrence count $C(v_{k_i}, v_{k_j})$, where $C(v_{k_i}, v_{k_j})$ denotes the total frequency of their joint sentence-level occurrences across all sentences.

\subsection{LLM-Driven Semantic Augmentation}

We employ an LLM-driven pipeline to enrich both content nodes and directory nodes with a new attribute $\mathrm{Summary}$, enabling effective relevance assessment during agentic retrieval.

\noindent\textbf{Content Node Summarization.} For each content node $v_c$, an LLM evaluates its textual content and produces a semantic summary capturing the core topic of the chunk. This concise summary enables rapid evaluation of chunk relevance during the agentic retrieval phase, e.g., content node selection in the directory backtracking procedure (Section 5.4), supplementing the raw content attribute with a high-level topical description.

\noindent\textbf{Directory Node Summarization.} Because directory nodes lack textual description, their semantic summaries are synthesized recursively from their immediate children in a bottom-up manner. Specifically, starting from the lowest directory level, the LLM takes a directory node's $\mathrm{Path}_{\mathrm{full}}$ alongside the summaries of all its immediate child nodes (including both sub-directories and content chunks) to generate a high-level domain summary. This process propagates upward level by level until reaching the root, producing hierarchical summaries that support directory-guided navigation and routing.

\subsection{Global Conflation}

After all nodes and edges from individual documents are instantiated and semantically enriched, we merge the per-document subgraphs into a unified global $\text{H}^2$KG through node deduplication. To be concrete, directory nodes are merged by matching their $\mathrm{Path}_{\mathrm{full}}$ and $\mathrm{Ver}_{\mathrm{prod}}$, ensuring that directories from different product versions are kept separate; content nodes are deduplicated using both \texttt{Content} and \texttt{Source} to prevent distinct chunks with identical text in different documents from being incorrectly merged; and entity nodes are deduplicated by matching their unique names. When entity nodes with the same name are merged, any duplicate co-occurrence edges between them are also merged by summing their weights into a single updated edge, and all directory-to-content edges referencing the removed node are redirected to the retained one.

\section{Agentic Retrieval of VDGR-RAG}

\subsection{Overview}

This section presents the agentic retrieval mechanism of VDGR-RAG, which coordinates a suite of composable tools to navigate the $\text{H}^2$KGs for accurate enterprise knowledge QA. As introduced in Section 1, the toolset consists of four components: a directory-enhanced routing tool, a multi-route retrieval tool, a directory backtracking tool, and a dynamic reflection tool. These components are dynamically orchestrated by the central LLM agent based on intermediate retrieval feedback to perform reasoning over the $\text{H}^2$KGs.

The rest of this section details each component. Section 5.2 presents knowledge base routing for directing queries to the most relevant domain-specific $\text{H}^2$KGs while filtering out cross-domain context noise. Section 5.3 details the multi-route retrieval tool that searches across multiple representation spaces (vector, directory, and graph) to maximize the comprehensiveness of knowledge recall. Section 5.4 presents directory backtracking for expanding the search scope across sibling branches when initial retrieval yields incomplete context. Section 5.5 describes the dynamic reflection tool that continuously evaluates the sufficiency and relevance of retrieved contexts and autonomously plans the next search query upon detecting information gaps. Finally, Section 5.6 introduces the agentic retrieval pipeline that composes these components into a unified process.

\subsection{Knowledge Base Routing}

\textbf{Problem Formulation.}
Given a user query $q$ and a collection of domain-specific $\text{H}^2$KGs $\{\mathcal{G}_1, \mathcal{G}_2, \ldots, \mathcal{G}_n\}$, the goal of knowledge base routing is to select a subset $\mathcal{G}^* \subseteq \{\mathcal{G}_1, \mathcal{G}_2, \ldots, \mathcal{G}_n\}$ that contains the required evidence for answering $q$, thereby filtering out irrelevant domains that introduce context noise. This step is particularly important when a query spans multiple technical domains with overlapping terminology, where routing not only reduces noise but also assists in identifying cross-domain knowledge.

\noindent\textbf{Directory-Node-Enhanced Routing.}
Our routing mechanism leverages both domain-specific knowledge and the hierarchical directory structure of each $\text{H}^2$KG to guide the search. For $\text{H}^2$KGs with deep directory hierarchies, only the directory nodes of the top-$N$ levels are preserved as the context for knowledge base routing, where $N$ is configurable based on application needs. The routing LLM analyzes the query against these directory nodes alongside domain knowledge to select target $\text{H}^2$KGs, assigning a confidence score to each candidate. $\text{H}^2$KGs with low confidence scores are excluded from the retrieval process to avoid introducing irrelevant context noise. The selected $\text{H}^2$KGs are then processed in descending order of confidence scores. For each $\text{H}^2$KG in this order, the retrieval pipeline (Section 5.6) is executed, and the LLM evaluates whether the currently gathered evidence is sufficient to answer the query. If sufficiency is confirmed, the pipeline terminates early and proceeds directly to answer synthesis, avoiding unnecessary retrieval over remaining $\text{H}^2$KGs.

\subsection{Multi-Route Retrieval}
\begin{figure*}[!t]
	\centering
	\includegraphics[width=\textwidth]{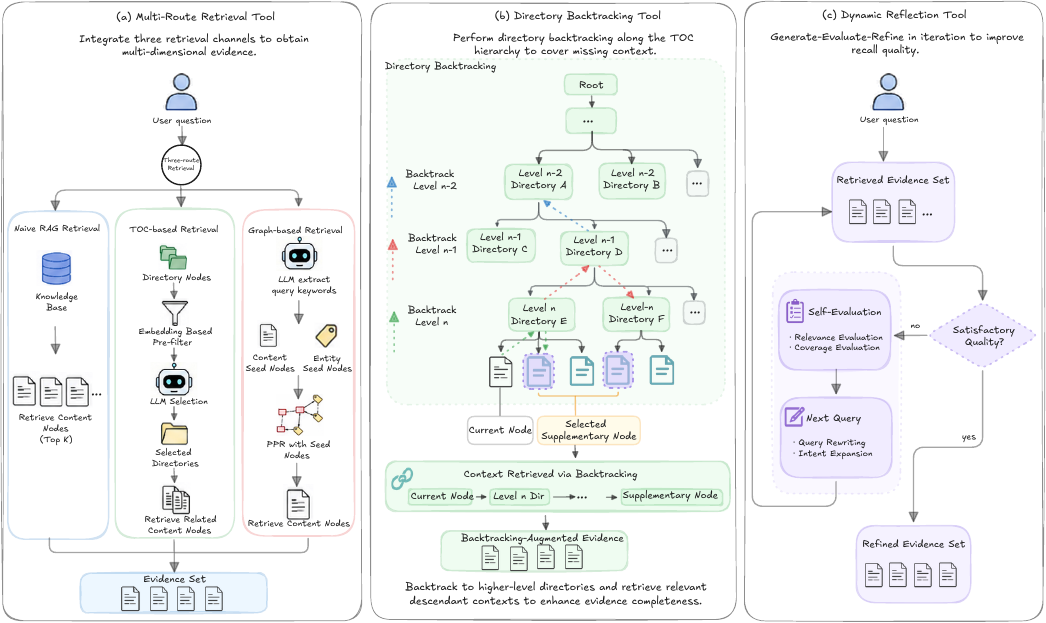}
	\caption{Overview of the agentic retrieval tools in VDGR-RAG. (a) Multi-route retrieval: three retrieval pathways (vector search, TOC-based agentic search, and entity-enhanced graph search) are executed and their results aggregated. (b) Directory backtracking: starting from the retrieved content nodes, the system traverses upward to parent directory nodes and expands to sibling branches to fetch missing prerequisite context. (c) Reflection tool: when retrieved evidence is insufficient, the LLM identifies knowledge gaps and generates a refined query for another round of retrieval.}
	\label{fig:retrieval-all_cropped}
\end{figure*}

Given the target $\text{H}^2$KG selected by knowledge base routing and a user query $q$, the system executes multiple independent retrieval routes. Each route explores a distinct representation space, and their results are combined during evidence aggregation, as shown in Figure~\ref{fig:retrieval-all_cropped}(a).

\noindent\textbf{Route 1: Naive RAG.}
Route 1 follows the naive RAG paradigm described in Section 3.3, combining sparse lexical search (BM25) and dense vector retrieval for hybrid candidate generation, followed by neural re-ranking. The output top-$K$ chunks are denoted as $\mathcal{C}_{\text{vec}}$.

\noindent\textbf{Route 2: TOC-Based Agentic Search.}
Route 2 leverages directory nodes in the $\text{H}^2$KG to retrieve structurally relevant context across three stages:
\begin{itemize}
	\item \textbf{Candidate Directory Nodes Selection.} The embedding model computes embeddings for the query $q$ and all directory nodes in $\mathcal{V}_\mathcal{D}$. Based on the computation of cosine similarity $\text{cosine}(\mathbf{e}_q, \mathbf{e}_d)$, the top-$L$ candidate directory nodes are retrieved as $\mathcal{D}_{\text{cand}}$, where $\mathbf{e}_q$ and $\mathbf{e}_d$ denote the embeddings of query $q$ and directory node $v_d$, respectively.
	\item \textbf{LLM Filtering.} An LLM evaluates all the candidate directories using their $\texttt{Path}_{\texttt{full}}$ and $\texttt{Summary}$ to prune irrelevant ones, producing a refined set of target directory nodes $\mathcal{D}_{\text{sel}}$. To favor specificity, the LLM is encouraged to prefer deeper directories from $\mathcal{D}_{\text{cand}}$, as deeper nodes correspond to more specific topics and yield a more focused retrieval scope.
	\item \textbf{Content Node Selection.} For each directory node $d \in \mathcal{D}_{\text{sel}}$, this route collects all content nodes reachable under the selected directories to form the retrieved chunk set $\mathcal{C}_{\text{dir}}$:
	\begin{equation}
		\mathcal{C}_{\text{dir}} = \bigcup_{d \in \mathcal{D}_{\text{sel}}} \{v \in \mathcal{V}_\mathcal{C} \mid \text{dir}(v) \in \text{subtree}(d)\}
	\end{equation}
	where $\text{dir}(v)$ denotes the parent directory node of content node $v$, and $\text{subtree}(d)$ denotes the set of all directory nodes in the subtree rooted at $d$.
\end{itemize}

\noindent\textbf{Route 3: Entity-Enhanced Graph Search.}
Route 3 leverages entity-entity and entity-content edges in the $\text{H}^2$KG to perform graph-based retrieval. This pipeline consists of four steps: query entity extraction, node score initialization, subgraph extraction, and score propagation.
\begin{itemize}
	\item \textbf{Query Entity Extraction.} The system extracts a set of domain-specific entities $\mathcal{Q}$ from query $q$.
	\item \textbf{Node Score Initialization.} Based on the extracted query entities, the system computes initial relevance scores for both content nodes and entity nodes:
	\begin{itemize}
		\item \textbf{Content Node Scores:} Each candidate content node $v_c$ is scored via multi-field keyword matching:
		\begin{equation}
	    s_{\text{content}}(c) = w_1 \cdot s_{\text{text}}(c) + w_2 \cdot s_{\text{title}}(c) + w_3 \cdot s_{\text{summary}}(c) + w_4 \cdot \sum_{\substack{v_k \in \mathcal{N}_\mathcal{K}(c) \\ \mathrm{Name}(v_k) \supseteq q_e, \; q_e \in \mathcal{Q}}} \lambda_{kc}
	    \end{equation}
	    where $s_{\text{text}}$, $s_{\text{title}}$, and $s_{\text{summary}}$ count the number of query entity hits from $\mathcal{Q}$ in the content node's textual content, title, and summary, respectively; $\mathcal{N}_\mathcal{K}(c)$ denotes the set of entity nodes connected to $v_c$; $\mathrm{Name}(v_k)$ denotes the name of entity node $v_k$, and $\lambda_{kc}$ denotes the importance score of entity node $v_k$ in content node $v_c$.
		\item \textbf{Entity Node Scores:} Each entity node $v_k$ is scored directly by its semantic similarity to the query:
		\begin{equation}
			s_{\text{entity}}(k) = \text{cosine}(\mathbf{e}_q, \mathbf{e}_k)
		\end{equation}
		where $\mathbf{e}_k$ denotes the embedding of the name of node $v_k$.
	\end{itemize}
	\item \textbf{Subgraph Extraction.} Based on the initial scores, we select a subgraph from the $\text{H}^2$KG for subsequent PPR propagation. For content nodes, we first identify the top-$N$ distinct scores and retain all content nodes whose scores fall within this set, thereby accommodating score ties; for entity nodes, we select the top-$M$ entity nodes. We then extract the subgraph induced by these selected nodes together with their first-order neighbors. 
	\item \textbf{Score Propagation.} We perform Personalized PageRank (PPR) to propagate relevance scores across the extracted subgraph:
	\begin{itemize}
        \item \textbf{Score Normalization:} To eliminate scale discrepancies between the two score types, we normalize each score within its respective candidate set:
	    \begin{align}
		    \hat{s}_{\text{content}}(c) &= \frac{s_{\text{content}}(c)}{\sum_{c' \in \mathcal{V}_\mathcal{C}} s_{\text{content}}(c')}, \nonumber \\
		    \hat{s}_{\text{entity}}(k) &= \frac{s_{\text{entity}}(k)}{\sum_{k' \in \mathcal{V}_\mathcal{K}} s_{\text{entity}}(k')}.
	    \end{align}
		\item \textbf{Personalization Vector Construction:} We construct the personalization vector $\mathbf{s}$ for the induced subgraph by combining the normalized scores with a balancing hyperparameter $\beta$:
		\begin{equation}
			\mathbf{s}(v) = 
			\begin{cases}
				\beta \cdot \hat{s}_{\text{content}}(v), & \text{if } v \in \mathcal{V}_\mathcal{C} \\[4pt]
				(1 - \beta) \cdot \hat{s}_{\text{entity}}(v), & \text{if } v \in \mathcal{V}_\mathcal{K} \\[4pt]
				0, & \text{otherwise}
			\end{cases}
		\end{equation}
        where $\mathbf{s}(v)$ denotes the personalization score assigned to node $v$, serving as the initial score for PPR, and $\beta\in[0,1]$ controls the relative contribution of content-node and entity-node scores.
		\item \textbf{PPR Propagation:} Using the personalization vector $\mathbf{s}$, PPR iteratively propagates relevance scores across the subgraph until convergence:
		\begin{equation}
			\boldsymbol{\pi}^{(t+1)} = (1 - \alpha) \mathbf{s} + \alpha \mathbf{A}^{\top} \mathbf{D}^{-1} \boldsymbol{\pi}^{(t)}
		\end{equation}
		where $\mathbf{A}$ denotes the adjacency matrix of the subgraph, which retains only co-occurrence and existence edges from the seed nodes and their one-hop neighbors; although existence edges are directed in the original graph, all edges are treated as undirected to allow bidirectional score propagation, with each edge weighted by $\log(1 + w)$ where $w$ is the co-occurrence count for co-occurrence edges and $w = 1$ for existence edges. $\mathbf{D}$ denotes the corresponding degree matrix, $\alpha$ is the damping factor of PPR, $t$ denotes the iteration step, and $\boldsymbol{\pi}^{(0)} = \mathbf{s}$ is initialized with the personalization vector.   
		\item \textbf{Content Node Selection:} The top-$K$ content nodes with the highest PPR scores after $T$ rounds of PPR propagation form the final graph retrieval set $\mathcal{C}_{\text{graph}}$.
	\end{itemize}
\end{itemize}

Upon collecting the content node sets $\mathcal{C}_{\text{vec}}$, $\mathcal{C}_{\text{dir}}$, and $\mathcal{C}_{\text{graph}}$ from all routes, we deduplicate the union and employ an LLM to extract knowledge relevant to the user query, forming the retrieved evidence set $\mathcal{E}$. By integrating vector search, TOC-based agentic search, and graph search, multi-route retrieval provides more comprehensive knowledge recall than prior approaches that focus on a single retrieval capability. Moreover, the three routes can be flexibly composed: they can be executed sequentially (e.g., starting with vector search and only invoking agentic or graph search if the LLM deems the retrieved evidence insufficient) or in parallel, as illustrated in Figure~\ref{fig:retrieval-all_cropped}(a).

\subsection{Directory Backtracking}
When the multi-route retrieval yields an insufficient evidence set $\mathcal{E}$, the retrieved results may exhibit localization bias relative to the true knowledge location. Since technical documentation usually groups related specifications under shared parent directories, the true knowledge may reside in the sibling nodes of the initially retrieved content nodes, or in the subtrees of their ancestor's siblings. Standard vector search often fails to cross these structural boundaries, and directory backtracking aims to correct such knowledge localization biases. The process of directory backtracking shown in Figure~\ref{fig:retrieval-all_cropped}(b) iterates as follows:

In the first round, the current nodes are all content nodes retrieved by Route 1, i.e., $\mathcal{C}_{\text{vec}}$. Their sibling nodes are also content nodes. The LLM first reads the summaries of these sibling content nodes to select relevant ones, and then extracts supplementary evidence from the selected content nodes to augment $\mathcal{E}$.

If the LLM judges the evidence still insufficient, the directory backtracking tool backtracks one level upward: the current nodes become the parent directory nodes of the previous round. Their sibling nodes are sibling directory nodes under the same grandparent directory. For every sibling directory, the LLM navigates top-down: it reads the summaries of the child nodes and selects the relevant ones to descend into, continuing until reaching the leaf content nodes. The LLM then selects relevant content nodes based on their summaries and extracts evidence from these nodes to augment $\mathcal{E}$.

This upward backtracking repeats until sufficient evidence is gathered or the maximum backtrack depth is reached, augmenting the evidence set $\mathcal{E}$ throughout the process.

\subsection{Dynamic Reflection}
When the retrieved evidence $\mathcal{E}$ is insufficient to support answering the user's query, the system can trigger the dynamic reflection tool to plan the next search query. 

Specifically, as shown in Figure~\ref{fig:retrieval-all_cropped}(c), the LLM self-evaluates the relevance and coverage of $\mathcal{E}$, identifies the current knowledge gaps, and generates a refined search query $q'$ for the next retrieval round:
\begin{equation}
	q' = \text{Reflect}(q, \mathcal{E})
\end{equation}
The process terminates when the model judges the evidence sufficient or the system reaches the configured maximum number of reflection iterations.

\subsection{Agentic Retrieval Pipeline}

\begin{algorithm}[t]
	\caption{VDGR-RAG Agentic Retrieval Pipeline}
	\label{alg:retrieval_pipeline}
	\begin{algorithmic}[1]
	\renewcommand{\algorithmicrequire}{\textbf{Input:}}
	\renewcommand{\algorithmicensure}{\textbf{Output:}}
	
	\Require Query $q$, Knowledge Base Collection $\{\mathcal{G}_1, \ldots, \mathcal{G}_n\}$, Enable Backtracking $B$, Enable Reflection $R$, Max Iterations $L_{\max}$
	\Ensure Synthesized Answer $a$
	
	\State $\mathcal{G}^* \leftarrow \text{KnowledgeBaseRouting}(q, \{\mathcal{G}_1, \ldots, \mathcal{G}_n\})$ \Comment{Route and rank by confidence}
	\State $\mathcal{E} \leftarrow \emptyset$, $\text{iter} \leftarrow 0$
	
	\Repeat
	\For{each $\mathcal{G}_i \in \mathcal{G}^*$ in descending confidence order}
	\State $\mathcal{E}_i \leftarrow \text{MultiRouteRetrieval}(q, \mathcal{G}_i)$ \Comment{Route 1, 2, 3}
	\State $\mathcal{E} \leftarrow \mathcal{E} \cup \mathcal{E}_i$
	
	\If{$B$ and $\text{not IsSufficient}(\mathcal{E}, q)$}
	\State $\mathcal{E} \leftarrow \text{DirectoryBacktrack}(\mathcal{C}_{\text{vec}}, \mathcal{G}_i) \cup \mathcal{E}$ \Comment{Backtrack from Route 1 content nodes}
	\EndIf
	
	\If{$\text{IsSufficient}(\mathcal{E}, q)$}
	\State \textbf{break} \Comment{Early exit when evidence sufficient}
	\EndIf
	\EndFor
	
	\If{$R$ and $\text{not IsSufficient}(\mathcal{E}, q)$}
	\State $q \leftarrow \text{DynamicReflection}(q, \mathcal{E})$
	\State $\text{iter} \leftarrow \text{iter} + 1$
	\EndIf
	\Until{$\text{IsSufficient}(\mathcal{E}, q)$ or $\text{iter} \geq L_{\max}$ or not $R$}
	
	\State $a \leftarrow \text{Generate}(q, \mathcal{E})$
	\State \textbf{return} $a$
	
\end{algorithmic}
\end{algorithm}

The components described above are fully modular and composable, enabling flexible pipeline configurations. Directory backtracking and dynamic reflection can each be independently enabled or disabled, allowing the system to balance retrieval quality and computational cost according to operational requirements.

The overall retrieval process proceeds as follows. First, knowledge base routing selects a subset of domain-specific $\text{H}^2$KGs $\mathcal{G}^*$ and ranks them by confidence score in descending order. For each $\text{H}^2$KG in this order, the pipeline executes multi-route retrieval to collect the initial evidence set $\mathcal{E}$; if $\mathcal{E}$ is already sufficient, the pipeline terminates early. Otherwise, if directory backtracking is enabled, the directory backtracking tool expands the search scope along the directory hierarchy to augment $\mathcal{E}$. If the evidence is still insufficient after all $\text{H}^2$KGs have been searched, and dynamic reflection is enabled, the dynamic reflection tool generates a refined query $q'$ and triggers an additional retrieval pass across all $\text{H}^2$KGs. The complete retrieval architecture is summarized in Algorithm~\ref{alg:retrieval_pipeline}.

\section{Experiments}
We conduct extensive experiments to evaluate VDGR-RAG on enterprise knowledge QA across four wireless telecommunications domains. We compare VDGR-RAG against representative baselines in terms of knowledge recall and answer accuracy, and perform ablation studies to validate the contribution of each retrieval tool and the knowledge base routing module.

\subsection{Setup}
\noindent\textbf{Datasets.} We evaluate our framework on enterprise product documents in the telecommunications domain across four datasets: Energy Saving (ES), Fault Management (FM), Experience Assurance (EA), and a large General Dataset (GD). The ES, FM, and EA datasets focus on specific sub-domains within telecommunications, while GD covers the general telecommunications domain. The test queries, provided by the company's internal testing team, cover simple single-hop retrieval, complex multi-hop reasoning, and cross-domain knowledge aggregation, totaling approximately 700 questions. Due to computational resource limits, we reproduce open-source baselines only on ES, FM, and EA; on the larger GD dataset, we conduct self-evaluation and ablation studies. Additionally, we combine the ES, FM, and EA datasets into a merged corpus to test the knowledge base routing module. The detailed statistics of these datasets are presented in Table~\ref{tab:dataset_specs}.

\noindent\textbf{Evaluation Metrics.}
We adopt two metrics to assess retrieval quality and end-to-end answer generation quality. Specifically, \textit{Retrieval Recall} (RR) measures the coverage of the ground-truth evidence by the retrieved context. \textit{Answer Accuracy} (AA) measures the accuracy of the generated answers against the ground-truth answers, reflecting the overall correctness of the RAG pipeline. Both metrics are evaluated by an LLM judge that compares each result against the ground truth, following the LLM-as-a-judge paradigm~\citep{zheng2023judging}.

\noindent\textbf{Baselines.}
We compare VDGR-RAG against five representative methods spanning two architectural paradigms. \textit{Graph-based RAG}: LinearRAG~\citep{2025LinearRAG} constructs a relation-free hierarchical graph (Tri-Graph) with lightweight entity extraction and semantic linking; E$^2$GraphRAG~\citep{2025E2GraphRAG} constructs a summary tree and an entity graph with bidirectional entity-chunk indexes for adaptive local-global retrieval; HippoRAG~\citep{Guti2024HippoRAG} applies Personalized PageRank (PPR) on an OpenIE knowledge graph for dynamic node ranking. \textit{Agentic RAG}: A-RAG~\citep{2026ARAG} exposes hierarchical retrieval interfaces (keyword search, semantic search, and chunk read) to the agent for adaptive multi-granularity retrieval; BookRAG~\citep{2025BookRAG} performs agentic planning over a hierarchical Tree-Graph BookIndex.

\noindent\textbf{Implementation Details.}
All methods share identical tokenization, embedding models (Qwen3-Embedding-8B)~\citep{2025Qwen3Embedding, 2025Qwen3}, and the same set of chunks produced by the company's internal document processing pipeline to ensure a fair comparison. The Qwen3.6-27B model serves as the backbone generator across all configurations. For VDGR-RAG, both directory backtracking and dynamic reflection are enabled by default, with the maximum backtrack depth and the maximum number of reflection iterations both set to 3. For multi-route retrieval, we set $K=10$ for Routes 1 and 3, $L=50$ for candidate directory node selection in Route 2, $N=10$ for the content node candidate selection in Route 3, $M=200$ for the entity node candidate selection in Route 3, and $\beta=0.8$ for the balancing hyperparameter in Route 3.

\subsection{Results}
\subsubsection{End-to-End QA Performance}
As shown in Table~\ref{tab:overall_result}, VDGR-RAG achieves the best performance across all three domain-specific datasets in both RR and AA. Notably, VDGR-RAG reaches 98.5\% RR / 97.6\% AA on ES and 98.7\% RR / 97.3\% AA on EA, substantially outperforming all baselines.

\begin{table}[t]
\centering
\caption{Dataset statistics. RR denotes Retrieval Recall, and AA denotes Answer Accuracy.}
\label{tab:dataset_specs}
\setlength{\tabcolsep}{8pt}
\begin{tabular}{lccccc}
\toprule
\textbf{Dataset} & \textbf{Alias}  & \textbf{\# of Tokens} & \textbf{\# of Chunks} & \textbf{\# of Entities} & \textbf{Metrics} \\ \midrule
Energy Saving        & ES   & 2764970 & 1927  & 6234 & RR, AA \\
Fault Management     & FM   & 38942   & 73     & 497  & RR, AA \\
Experience Assurance & EA   & 448716  & 333    & 1747 & RR, AA \\
General Dataset      & GD   & 72916106 & 64823  & 91846 & RR, AA \\
\bottomrule
\end{tabular}
\end{table}

\begin{table*}[!t]
	\centering
    \setlength{\tabcolsep}{6pt}
    \small
	\caption{Performance comparison of baseline methods and the proposed framework across domain-specific QA datasets. The best results are highlighted in \textbf{bold}.}
	\label{tab:overall_result}
	\begin{tabular}{llccccccc}
		\toprule
		\textbf{Paradigm} & \textbf{Method} & \multicolumn{2}{c}{\textbf{ES}} & \multicolumn{2}{c}{\textbf{FM}} & \multicolumn{2}{c}{\textbf{EA}} \\
		\cmidrule(lr){3-4} \cmidrule(lr){5-6} \cmidrule(lr){7-8}
		& & RR (\%) & AA (\%) & RR (\%) & AA (\%) & RR (\%) & AA (\%) \\ \midrule
		\multirow{3}{*}{\textbf{Graph-based RAG}} & LinearRAG & 70.3 & 55.4 & 88.1 & 75.3 & 95.5 & 85.6 \\
		& E$^2$GraphRAG & 75.6 & 60.2 & 86.8 & 75.9 & 85.7 & 77.2 \\
		& HippoRAG & 82.3 & 77.1 & 88.1 & 69.6 & 81.7 & 85.5 \\ \midrule
		\multirow{2}{*}{\textbf{Agentic RAG}} & A-RAG & 66.3 & 63.3 & 80.1 & 83.9 & 81.0 & 84.1 \\
		& BookRAG & 84.8 & 74.9 & 91.9 & 83.2 & 90.9 & 89.7 \\ \midrule
		\multirow{1}{*}{\textbf{Ours}} & VDGR-RAG & \textbf{98.5} & \textbf{97.6} & \textbf{92.4} & \textbf{91.7} & \textbf{98.7} & \textbf{97.3} \\
		\bottomrule
	\end{tabular}
\end{table*}

\noindent\textbf{Comparison with Graph-based RAG.}
Among the three graph-based RAG baselines, both LinearRAG and E$^2$GraphRAG rely on spaCy for entity extraction, which performs poorly on telecommunications domain documents due to the prevalence of specialized terminology and product-specific identifiers. This leads to low-quality graphs with missing or incorrect entities, resulting in poor AA (e.g., 55.4\% and 60.2\% on ES). HippoRAG relies solely on PPR-based graph retrieval, which limits its ability to capture hierarchical document structures and causes poor performance. In contrast, VDGR-RAG leverages an LLM for entity extraction and constructs the $\text{H}^2$KG to preserve both directory structures and semantic relationships, enabling precise knowledge localization.

\noindent\textbf{Comparison with Agentic RAG.}
A-RAG achieves the best AA on the FM dataset among all the baselines through adaptive multi-granularity search. However, without leveraging document structure, A-RAG relies solely on keyword matching and sentence-level semantic similarity, which is insufficient for efficient and accurate agentic search path selection in hierarchical technical documentation. BookRAG achieves relatively strong performance among the baselines (e.g., 90.9\% RR / 89.7\% AA on EA) by providing both graph-based and TOC-based reasoning retrieval; however, these two retrieval modes are tightly coupled within a single retrieval path, where the role of TOC-based reasoning is to select entity seed nodes for graph PPR. This coupling design causes the two modes to interfere with each other, preventing the complementary strengths of both modes from being fully exploited. VDGR-RAG addresses these limitations through decoupled multi-route retrieval, where each route independently targets a distinct representation space—vector similarity, directory structure, and entity topology—while their results are aggregated at the evidence level.

\noindent\textbf{Effectiveness of Multi-Route Retrieval.}
Comparing across paradigms, graph-based RAG and agentic RAG each excel on different datasets. For instance, on ES, HippoRAG achieves the highest AA (77.1\%), while on FM, A-RAG attains the best AA (83.9\%). This cross-paradigm variation suggests that both graph-based and agentic retrieval are beneficial. BookRAG, which employs Tree-Graph-based agentic retrieval, achieves the best overall performance among all baselines across the three datasets, confirming the complementarity and necessity of integrating graph-based and agentic retrieval modes. However, as discussed above, BookRAG couples graph search and TOC-based reasoning within a single retrieval path, preventing each mode's individual strengths from being fully leveraged. VDGR-RAG substantially outperforms BookRAG across all datasets, demonstrating that decoupled multi-route retrieval with unified reasoning effectively leverages the complementary strengths of each route.

\subsubsection{Ablation Studies and Analysis}

\noindent\textbf{Ablation Study 1: Retrieval Tool Effectiveness.}
Table~\ref{tab:ablation_retrieval} presents a progressive ablation study on the GD dataset. Starting from the naive RAG baseline (M1), we incrementally add each retrieval component and observe its marginal contribution. Key findings are as follows:

\begin{table}[t] 
	\centering
	\caption{Progressive ablation of retrieval tools on GD.}
	\label{tab:ablation_retrieval}
	\begin{tabular}{clcc}
		\toprule
		\textbf{ID} & \textbf{Configuration} & \textbf{RR (\%)} & \textbf{AA (\%)} \\ \midrule
		\textbf{M1} & Naive RAG Baseline & 80.5 & 76.5 \\
		\textbf{M2} & M1 + TOC-Based Agentic Search & 91.3 & 88.6 \\
		\textbf{M3} & M2 + Entity-Enhanced Graph Search & 92.0 & 90.5 \\ 
		\textbf{M4} & M2 + Dynamic Reflection & 94.5 & 92.4 \\
		\textbf{M5} & M3 + Dynamic Reflection & 96.9 & 95.7 \\
		\bottomrule
	\end{tabular}
\end{table}

\begin{table}[t]
	\centering
	\caption{Effect of directory backtracking on ES.}
	\label{tab:ablation_backtrack}
	\begin{tabular}{clcc}
		\toprule
		\textbf{ID} & \textbf{Configuration} & \textbf{RR (\%)} & \textbf{AA (\%)} \\ \midrule
		\textbf{B1} & Multi-Route Retrieval & 97.2 & 95.8 \\
		\textbf{B2} & B1 + Directory Backtracking & 98.3 & 97.0 \\
		\bottomrule
	\end{tabular}
\end{table}

\begin{itemize}
	\item \textbf{Naive RAG provides a strong and efficient baseline.} M1 already achieves 80.5\% RR and 76.5\% AA, demonstrating that hybrid sparse-dense retrieval with neural re-ranking can capture a substantial portion of relevant evidence. Given its low computational cost and minimal latency, naive RAG serves as an indispensable foundation in the RAG system, and all subsequent improvements are built upon this efficient baseline.
	\item \textbf{TOC-based agentic search drives the primary gain.} Adding TOC-based agentic search ($\text{M1}\!\rightarrow\!\text{M2}$) yields the largest improvement of 12.1\% AA. This confirms that leveraging hierarchical document structure is essential for domain-specific QA: TOC-based agentic search allows the LLM to directly locate knowledge through directory-driven reasoning rather than relying on embedding similarity, thereby overcoming the fundamental limitation that semantic similarity does not equate to relevance.
	\item \textbf{Graph search and reflection provide complementary benefits.} From the M2 baseline, adding entity-enhanced graph search ($\text{M2}\!\rightarrow\!\text{M3}$) improves AA by 1.9\%, while adding dynamic reflection ($\text{M2}\!\rightarrow\!\text{M4}$) improves AA by 3.8\%. These two modules address different failure modes: graph search discovers cross-chunk entity relationships that flat retrieval misses, thereby facilitating multi-hop reasoning, while dynamic reflection iteratively evaluates evidence sufficiency and generates refined queries to fill remaining knowledge gaps.
	\item \textbf{Full integration yields the best results.} Combining all components (M5) achieves 96.9\% RR and 95.7\% AA, a 19.2\% improvement over the naive RAG baseline. The fact that M5 outperforms both M3 and M4 indicates that graph search and dynamic reflection are mutually reinforcing rather than redundant: graph search expands the evidence boundary, while reflection refines the retrieval focus, forming a self-improving retrieval cycle. These results confirm that vectors, directories, graphs, and reflection are all necessary—each component addresses a distinct failure mode, and their synergy is essential for achieving the best performance.
    \item \textbf{Effect of directory backtracking.} As shown in Table~\ref{tab:ablation_backtrack}, on the ES dataset, adding directory backtracking to multi-route retrieval improves RR from 97.2\% to 98.3\% and AA from 95.8\% to 97.0\%, demonstrating that backtracking can recover relevant content missed by initial retrieval through navigating along the directory tree. However, on the GD dataset, the directory structure is extremely large and deeply nested. Backtracking with a maximum depth of 3 provides negligible gains, as the relevant content often resides more than 3 levels deep in the hierarchy. This suggests that directory backtracking is most beneficial when the document hierarchy is moderate in scale.
\end{itemize}

\noindent\textbf{Ablation Study 2: Knowledge Base Routing.}
Table~\ref{tab:routing_ablation} evaluates the routing module on the merged ES+FM+EA corpus. Without routing, the agent searches all domains jointly, introducing cross-domain entity noise and misleading terminology. Knowledge base routing eliminates this contamination by directing queries to the target domain $\text{H}^2$KG, improving RR by 4.2\% and AA by 2.8\%. Moreover, knowledge base routing is also expected to reduce retrieval latency by restricting the search space to the target domains, making it critical for large-scale deployments.

\begin{table}[!t] 
\centering
\caption{Ablation of knowledge base routing on the merged ES+FM+EA corpus.}
\label{tab:routing_ablation}
\begin{tabular}{lcc}
\toprule
\textbf{Method} & \textbf{RR (\%)} & \textbf{AA (\%)} \\ \midrule
VDGR-RAG w/o Routing & 92.4 & 92.3 \\
VDGR-RAG w/ Routing & 96.6 & 95.1 \\
\bottomrule
\end{tabular}
\end{table}

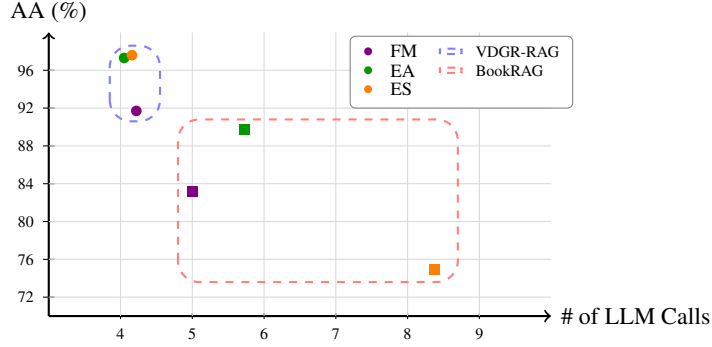
\begin{figure}[t]
	\centering
	\begin{tikzpicture}[xscale=0.95, yscale=0.5]
		\draw[->, thick] (0,0) -- (7,0) node[right] {\small \# of LLM Calls};
		\draw[->, thick] (0,0) -- (0,7.5) node[above] {\small AA (\%)};
		\foreach \x/\label in {1/4, 2/5, 3/6, 4/7, 5/8, 6/9} {
			\draw (\x,0.08) -- (\x,-0.08) node[below] {\tiny \label};
		}
		\foreach \y/\label in {0.5/72, 1.5/76, 2.5/80, 3.5/84, 4.5/88, 5.5/92, 6.5/96} {
			\draw (0.08,\y) -- (-0.08,\y) node[left] {\tiny \label};
		}
		\foreach \x in {1,...,6} {
			\draw[gray!30] (\x,0) -- (\x,7.5);
		}
		\foreach \y in {0.5,1.5,2.5,3.5,4.5,5.5,6.5} {
			\draw[gray!30] (0,\y) -- (7,\y);
		}
		\node[draw=none, fill=violet, rectangle, minimum size=4pt, inner sep=0pt] at (2.0, 3.3) {};
		\node[draw=none, fill=green!60!black, rectangle, minimum size=4pt, inner sep=0pt] at (2.73, 4.925) {};
		\node[draw=none, fill=orange, rectangle, minimum size=4pt, inner sep=0pt] at (5.37, 1.225) {};
		\node[draw=none, fill=violet, circle, minimum size=4pt, inner sep=0pt] at (1.22, 5.425) {};
		\node[draw=none, fill=green!60!black, circle, minimum size=4pt, inner sep=0pt] at (1.05, 6.825) {};
		\node[draw=none, fill=orange, circle, minimum size=4pt, inner sep=0pt] at (1.16, 6.9) {};
		\draw[blue!50, dashed, thick, rounded corners=8pt] (0.85, 5.15) rectangle (1.55, 7.15);
		\draw[red!50, dashed, thick, rounded corners=8pt] (1.8, 0.9) rectangle (5.7, 5.2);
		\draw[rounded corners=2pt, fill=white, draw=gray] (4.2, 5.5) rectangle (7.3, 7.4);
		\node[draw=none, fill=violet, circle, minimum size=3pt, inner sep=0pt] at (4.45, 7.0) {};
		\node[font=\scriptsize, right] at (4.63, 7.0) {FM};
		\draw[blue!50, dashed, thick, rounded corners=2pt] (5.45, 6.93) rectangle (5.75, 7.07);
		\node[font=\scriptsize, right] at (5.82, 7.0) {\tiny VDGR-RAG};
		\node[draw=none, fill=green!60!black, circle, minimum size=3pt, inner sep=0pt] at (4.45, 6.5) {};
		\node[font=\scriptsize, right] at (4.63, 6.5) {EA};
		\draw[red!50, dashed, thick, rounded corners=2pt] (5.45, 6.43) rectangle (5.75, 6.57);
		\node[font=\scriptsize, right] at (5.82, 6.5) {\tiny BookRAG};
		\node[draw=none, fill=orange, circle, minimum size=3pt, inner sep=0pt] at (4.45, 6.0) {};
		\node[font=\scriptsize, right] at (4.63, 6.0) {ES};
	\end{tikzpicture}
	\caption{Comparison of Answer Accuracy (AA) vs.\ LLM calls per query between VDGR-RAG and BookRAG across three datasets. Each dataset is shown in a distinct color. Square markers denote BookRAG and circle markers denote VDGR-RAG.}
	\label{fig:efficiency}
\end{figure}

\subsection{Efficiency Analysis}

Beyond retrieval quality, the computational cost of agentic retrieval is a critical concern for practical deployment. We compare the LLM invocation efficiency of VDGR-RAG against BookRAG, the strongest baseline that also adopts a similar graph structure combining a directory tree with an entity graph, as the number of LLM calls per query directly determines both latency and cost.

As shown in Figure~\ref{fig:efficiency}, VDGR-RAG achieves significantly higher accuracy while requiring far fewer LLM calls. On average, VDGR-RAG requires 4.14 LLM calls per query across the three datasets, compared to 7.22 for BookRAG—a 42.7\% reduction. This efficiency gap is most pronounced on ES, where BookRAG requires 8.37 calls per query while VDGR-RAG needs only 4.16, yet achieves 97.6\% AA versus 74.9\%. The key reason is that BookRAG first determines whether a query is complex and decomposes complex queries into sub-questions, then independently performs retrieval for each sub-question, leading to multiple LLM calls proportional to the number of sub-questions. In contrast, VDGR-RAG relies on more robust multi-route retrieval that directly addresses the original query without decomposition, thereby reducing the number of LLM invocations.

\section{Conclusion}
In this paper, we present VDGR-RAG, an agentic GraphRAG framework that integrates vectors, directories, graphs, and reflection for accurate enterprise knowledge QA. At the core of VDGR-RAG lies the $\text{H}^2$KG, a hierarchical heterogeneous knowledge graph that preserves both document directory structures and entity relationships. Over the $\text{H}^2$KG, VDGR-RAG orchestrates several composable retrieval tools: a directory-node-enhanced routing tool that directs queries to the target domain knowledge base, a multi-route retrieval tool that combines vector search, TOC-based agentic search, and entity-enhanced graph search for comprehensive evidence collection, a directory backtracking tool that corrects knowledge localization biases, and a dynamic reflection tool that iteratively refines the retrieval process. Extensive experiments on enterprise product documents across four telecommunications domains demonstrate that VDGR-RAG significantly outperforms representative baselines in terms of both retrieval recall and answer accuracy.

\bibliographystyle{plainnat}
\bibliography{references}

\end{document}